\documentclass[]{fairmeta}
\usepackage{makecell}
\usepackage{wrapfig}
\usepackage{tabularx}
\usepackage{textcomp}
\usepackage{stfloats}
\usepackage{url}
\usepackage{verbatim}
\usepackage{titlesec}
\usepackage{tocloft}
\usepackage{adjustbox}
\usepackage{multirow}
\usepackage{pifont}
\usepackage[sc]{mathpazo}
\usepackage{tikz}
\usepackage{comment}
\usepackage{amsmath,amssymb}
\usepackage{colortbl}
\usepackage{natbib}
\usepackage{color}
\usepackage{booktabs} 
\usepackage{hyperref}
\usepackage{graphicx}
\usepackage{subcaption}
\RequirePackage{xspace}
\makeatletter
\DeclareRobustCommand\onedot{\futurelet\@let@token\@onedot}
\def\@onedot{\ifx\@let@token.\else.\null\fi\xspace}
\usepackage[most]{tcolorbox}
\usepackage{array}
\usepackage{siunitx}
\usepackage[table]{xcolor}
\usepackage{caption}
\definecolor{headerpurple}{HTML}{d8d2fc}
\definecolor{rowgray}{gray}{0.95}
\usepackage{CJKutf8}

\makeatother

\definecolor{adptorange}{RGB}{248, 205, 172}
\definecolor{cmpblue}{RGB}{189, 215, 238}

\definecolor{our_red}{RGB}{232,157,160}
\definecolor{our_blue}{RGB}{136,206,230}
\definecolor{our_orange}{RGB}{246,200,168}
\definecolor{our_green}{RGB}{178,211,164}

\definecolor{attn_code0}{RGB}{247,215,200}
\definecolor{attn_code1}{RGB}{238,169,139}
\definecolor{mlp_code0}{RGB}{204,201,221}
\definecolor{mlp_code1}{RGB}{102,95,153}
\definecolor{mygray}{HTML}{f0f0f0}

\definecolor{token_blue}{RGB}{84, 120, 140}

\usepackage{bbding}
\usepackage{fontawesome}
\usepackage{float}

\newlength\savewidth

\newcolumntype{x}[1]{>{\centering\arraybackslash}p{#1pt}}
\newcolumntype{y}[1]{>{\raggedright\arraybackslash}p{#1pt}}
\newcolumntype{z}[1]{>{\raggedleft\arraybackslash}p{#1pt}}

\renewcommand{\paragraph}[1]{\vspace{1.25mm}\noindent\textbf{#1}}

\usepackage{algorithm}
\usepackage{listings}

\definecolor{codeblue}{rgb}{0.25, 0.5, 0.5}
\definecolor{codekw}{rgb}{0.35, 0.35, 0.75}
\lstdefinestyle{Pytorch}{
    language = Python,
    backgroundcolor = \color{white},
    basicstyle = \fontsize{9pt}{8pt}\selectfont\ttfamily\bfseries,
    columns = fullflexible,
    aboveskip=1pt,
    belowskip=1pt,
    breaklines = true,
    captionpos = b,
    commentstyle = \color{codeblue},
    keywordstyle = \color{codekw},
}

\definecolor{green}{HTML}{009000}
\definecolor{red}{HTML}{ea4335}

\title{WorldRuler: Interactive World Modeling under Configurable Rules}

\author[1,2]{Mingliang Zhai*}
\author[1]{Chuanhao Li$\dagger$}
\author[1,2]{Jiaming Tan}
\author[1,4]{Zhen Li}
\author[2]{Yuwei Wu}
\author[5]{Yunde Jia}
\author[3]{Zian Meng}
\author[3]{Yuanyang Yin}
\author[1,3]{Kaipeng Zhang}

\affiliation[1]{Alaya Lab}
\affiliation[2]{Beijing Institute of Technology}
\affiliation[3]{Shanghai Innovation Institute}
\affiliation[4]{The University of Tokyo}
\affiliation[5]{Shenzhen MSU-BIT University}

\newcommand{\modelname}{WorldRuler}
\newcommand{\dataname}{RuleVerse}

\abstract{

Rules govern how player actions change interactive worlds.
Different rules create distinct gameplay experiences within similar scenes and action spaces.
For customizable gameplay, generative world models must follow user-defined rules beyond real-world behaviors.
However, realizing user-defined rules presents two challenges: \textbf{Rule-State Persistence} and \textbf{Rule-to-Visual Realization}. 
The former requires preserving rule definitions, dependencies, and world states across saving and loading, so that restored worlds follow the established rules. 
The latter requires accurately realizing rule-induced changes on designated objects and attributes, particularly when affected objects are unseen.
In this paper, we propose \modelname{}, which organizes rule definitions, dependencies, and object states in an executable 3D rule graph and connects rule execution, video generation, and perceptual feedback in a closed loop. 
The graph supports joint saving and restoration of rules and world states, while dynamic object binding grounds rule-specified entities in visual instances and uses stored states to constrain generation from their first appearance. 
We further introduce object attribute state replacement and  counterfactual reconstruction training, enabling accurate attribute control and compatible appearance and motion changes while limiting interference with unrelated content.
We develop an automated data collection and annotation pipeline using GTA-V and Habitat to construct \dataname{}, a dataset of rule-driven interactions with detailed state and event annotations, including cases where rules take effect before affected objects become visible.
Extensive experiments demonstrate the effectiveness of \modelname{} in preserving rule-state consistency across saving and loading and accurately realizing rule outcomes on designated objects and attributes.

}

\github{\url{https://alaya-lab.github.io/\modelname}}
\Code{\url{https://github.com/AlayaLab/\modelname}}
\Data{\url{https://huggingface.co/datasets/AlayaLab/\dataname}}

\contact{chuanhao.li@shanda.com}
\date{\today}

\begin{document}
\maketitle

\begingroup
\renewcommand{\thefootnote}{}
\footnotetext{* Work done during internship at Alaya Lab, $\dagger$ {Corresponding author}}
\endgroup

\begin{figure*}[!h]
    \centering
    \includegraphics[width=\linewidth]{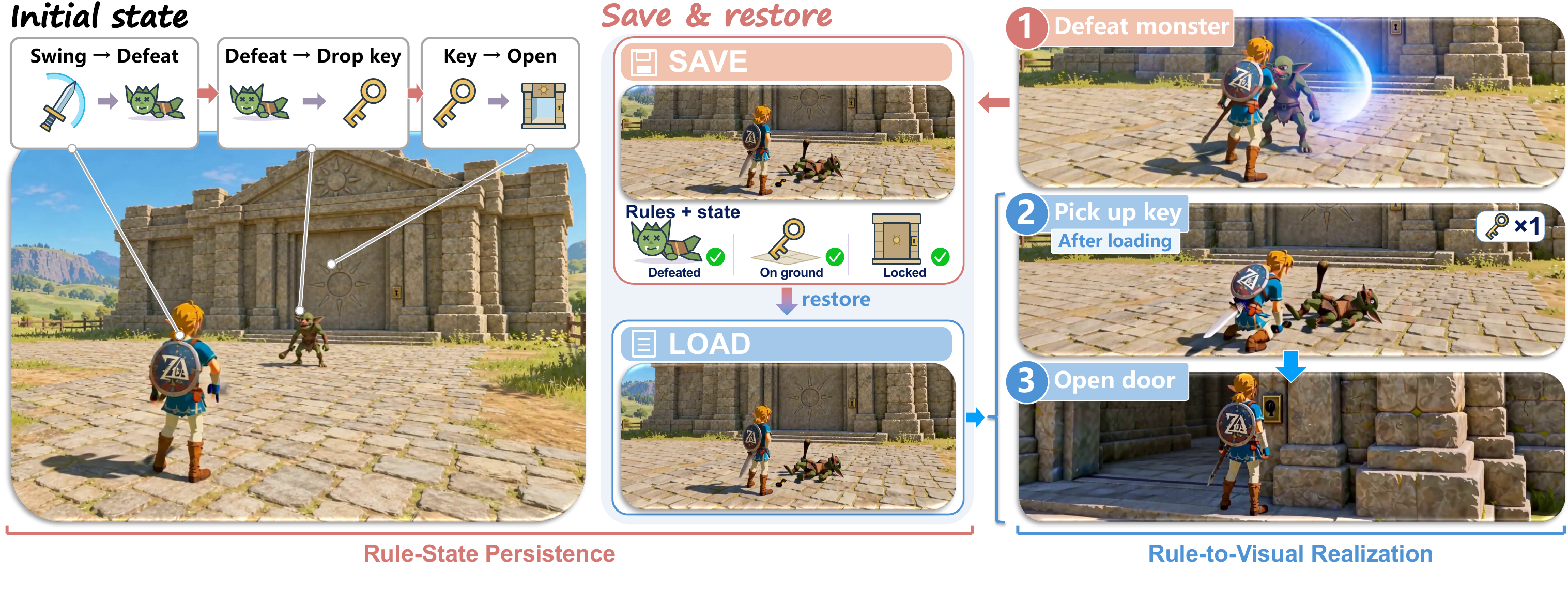}
    \caption{
    \textbf{WorldRuler enables persistent, rule-governed generative worlds.} 
    The saved rules and world state allow the player to collect the key and open the door after loading, demonstrating \textbf{Rule-State Persistence} and \textbf{Rule-to-Visual Realization}.
    }
    \label{fig:teaser}
\end{figure*}

\section{Introduction}
\label{sec:introduction}

Rules govern the consequences of player actions, creating different gameplay goals and strategies even within similar scenes and action spaces~\cite{hunicke2004mda}.
For example, \texttt{Cat Mario} subverts conventional game rules, where seemingly safe platforms collapse and power-ups can kill players, while \texttt{It Takes Two} introduces unique interaction rules, such as using nails to create paths across otherwise impassable gaps. 
These examples motivate generative game worlds~\cite{he2025matrixgame20,chen2026h3world} that consistently follow user-defined rules beyond conventional real-world behaviors.

Achieving this goal requires connecting user-defined interaction logic with world-state maintenance and visual generation. 
Code World Model~\cite{chen2026codeworldmodelcoding} and Programmable World Model~\cite{huang2026programmableworldmodel} maintain executable world states and translate them into video generation conditions.
Supporting persistent interaction requires both \textbf{Rule-State Persistence} across saving and loading and \textbf{Rule-to-Visual Realization} on designated visual instances.

Rule-State Persistence requires jointly preserving rule definitions, dependencies, and world states across saving and loading.
As shown in Figure~\ref{fig:teaser}, consider a rule chain in which a sword attack defeats a monster, the monster drops a key, and the key unlocks a door.
If the player saves before collecting the key, restoration must preserve the defeated monster, dropped key, closed door, and remaining rules.
A restorable world must therefore retain the states and rule dependencies that govern future interactions, including information absent from the current view.

Rule-to-Visual Realization requires grounding rule-specified objects and attributes in visual instances to realize changes at the correct time.
This is particularly important when the affected object is initially unseen~\cite{duan2026liveworldsimulatingoutofsightdynamics,gu2026worldcraftcameranavigationobject}.
For example, if a switch opens a gate outside the current view, the gate should already be open when the player first sees it.
The stored state must constrain this first appearance and remain effective in subsequent interactions.
Graph representations provide structured conditions for object relations and interactions in video generation~\cite{shah2026graphvid,park2026bringing}.
Custom rules further require persistent object binding and attribute control that supports compatible appearance and motion changes without disrupting unrelated content~\cite{li2025magicmotioncontrollablevideogeneration,chu2025wanmovemotioncontrollablevideogeneration}.

In this paper, we propose \modelname{}, which connects rule execution, video generation, and perceptual feedback through an executable 3D rule graph. 
The graph organizes rule definitions, dependencies, object identities, and attribute states together with spatial information. Rules and world states are saved and restored jointly, allowing subsequent actions to be evaluated against the restored conditions. 
Dynamic object binding associates rule-specified entities with visual instances and uses stored states to constrain generation from their first appearance. 
Perceptual feedback verifies generated outcomes and updates object correspondences and local geometry for subsequent binding and generation, while keeping rule-prescribed states separate from visual evidence.

To enforce these rule-prescribed states, we introduce an object attribute state replacement mechanism that reads object attribute states during denoising and replaces them according to the prescribed targets, correcting deviations in the controlled attributes.
We complement this mechanism with counterfactual reconstruction training to produce coherent visual outcomes.
Counterfactual reconstruction learns the visual realization of each state, while paired supervision encourages compatible changes and preservation of unrelated content.
Training constrains these visual adjustments to remain compatible with the prescribed states, supporting the required appearance and motion changes while limiting interference with other objects and attributes.

To support counterfactual reconstruction training and rule consistency evaluation, we develop an automated data collection and annotation pipeline using GTA-V and Habitat~\cite{puig2024habitat}, resulting in \textbf{\dataname{}}, a dataset of rule-driven interactions with detailed annotations of object states, trigger conditions, rule consequences, and event timing. 
\dataname{} covers diverse triggers and state changes, multi-step rule dependencies, and rules that take effect before the affected objects become visible. 
\dataname{} supports the evaluation of rule-state consistency across saving and loading, the accurate realization of prescribed outcomes on designated objects and attributes, and preservation of unrelated content.
Experiments on the public WorldState-Bench~\cite{ki2026stateworldmodelsexecute} and our \dataname{} demonstrate the effectiveness of \modelname{}.
In summary, our contributions are threefold:

\begin{itemize}
    \item We propose \modelname{}, an executable 3D rule graph framework with dynamic object binding and perceptual feedback for rule persistence and visual realization.
    
    \item We introduce object attribute state replacement with counterfactual reconstruction training to realize prescribed changes while preserving unrelated content.
    
    \item We construct \dataname{} through automated data collection and annotation in GTA-V and Habitat, supporting counterfactual training and rule consistency evaluation.
\end{itemize}

\section{Related Work}
\label{sec:related_work}

\subsection{Video World Models}

Video world models predict observations from visual history and actions for exploration~\cite{kim2020gamegan,yang2024unisim}.
Early methods learn environment dynamics and action-conditioned generation~\cite{bruce2024genie,alonso2024diamond,valevski2025gamengen}.
Recent work advances real-time control and long-horizon consistency.
Yume1.5~\cite{mao2026yume15} combines context compression and attention distillation for real-time generation with text-controlled events.
Long interactions, however, introduce inconsistencies when revisiting scenes.
WorldPlay~\cite{sun2025worldplay} improves geometric consistency through reconstituted context memory.
Matrix-Game 3.0~\cite{wang2026matrixgame30} combines camera-aware memory retrieval and self-correction training.
AlayaWorld~\cite{team2026alayaworldv11} uses a streaming 3D point cache for spatiotemporal consistency.
AlayaVista~\cite{tan2026alayavista} uses panoramic latent states and viewport rendering for perspective generation.
Alaya-EVOKE~\cite{yin2026alayaevoke} combines external spatial memory with long-horizon supervision.
Despite these advances, persistent adherence to user-defined rules remains insufficiently addressed.
WorldRuler maintains rule states to constrain subsequent video generation.

\subsection{Controllable Video Generation}

Controllable video generation aims to realize specified changes while preserving unrelated content.
Early methods use camera poses and motion trajectories for generation control~\cite{wang2024motionctrl,he2024cameractrl,yin2023dragnuwa}.
H3-World~\cite{chen2026h3world} translates language instructions into aligned character and camera controls.
More complex object and structural changes require explicit representations.
WorldCraft~\cite{gu2026worldcraftcameranavigationobject} uses 3D representations and world-coordinate constraints for object manipulation.
CustomX~\cite{wang2025customx} customizes characters, actions, and scenes using 3D assets.
SolarRenderer~\cite{fang2026solarrenderer} adapts pretrained video generators to structured scene representations.
GraphVid~\cite{shah2026graphvid} uses interaction graphs for multi-entity control.
Park et al.~\cite{park2026bringing} introduce graph-aligned denoising supervision through paired video-graph modifications.
However, these methods do not explicitly maintain rule-dependent states across interactions.
WorldRuler binds executable rule states to persistent visual instances and uses attribute replacement to realize prescribed changes on designated objects while preserving unrelated visual content.

\subsection{State and Rules in World Models}

Persistent states maintain interaction consequences beyond current observations.
Early world models learn latent dynamics~\cite{ha2018worldmodels,hafner2025dreamerv3}, and structured representations capture object identities and spatial relationships~\cite{wu2023slotformer}.
However, unobserved states cannot always be recovered from visual history.
LiveWorld~\cite{duan2026liveworldsimulatingoutofsightdynamics} models dynamic objects outside the current view.
Executable logic enables explicit control over state transitions.
Code World Model~\cite{chen2026codeworldmodelcoding} uses a coding agent to maintain world states and compile them into proxy videos.
Programmable World Model~\cite{huang2026programmableworldmodel} maintains global states through executable programs and converts them into spatial generation conditions.
State World Models~\cite{ki2026stateworldmodelsexecute} execute interaction rules and updates states using generated observations.
Building on explicit state maintenance and rule execution, WorldRuler jointly restores rule definitions, dependencies, and execution states, ensuring their consequences are realized on designated visual instances.

\section{\modelname}
\label{sec:method}

\subsection{Problem Formulation and Overview}
\label{sec:method-overview}

Given an initial visual observation $I_0$, user-defined rules $\mathcal{R}$, and a sequence of player actions, WorldRuler generates interactive videos that follow these rules. 
An executable 3D rule graph $\mathcal{G}_t$ maintains rule definitions and dependencies, an execution state $X_t$ records object attributes and rule progress, and bindings $\mathcal{B}_t$ associate rule entities with 3D objects and visual instances. 
Together, these representations determine subsequent interactions and persist when objects leave the view or a session is saved.

\begin{figure*}[t]
    \centering
    \includegraphics[width=\textwidth]{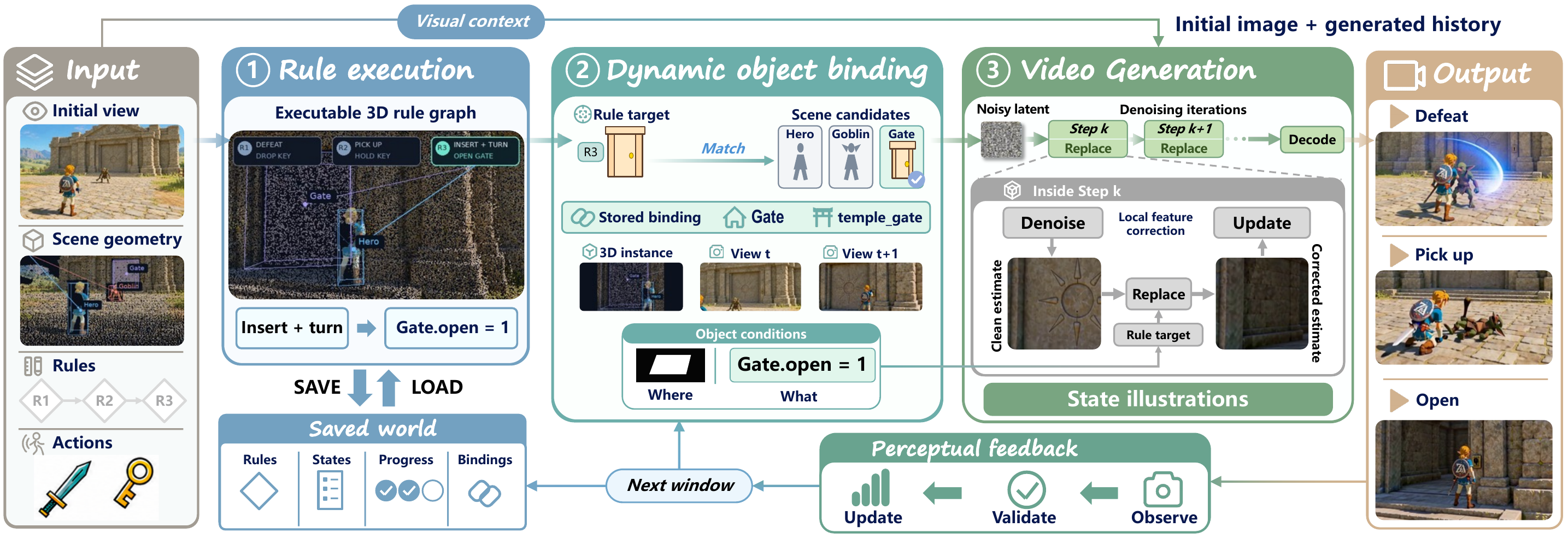}
    \caption{
    Overview of WorldRuler. 
    Rule execution updates persistent object states, dynamic binding associates rule entities with 3D instances and their current image regions, and local attribute replacement corrects denoising estimates for the next sampling update. 
    Rules, states, execution progress, and bindings are saved and restored jointly. 
    Independent perceptual feedback informs subsequent windows. 
    Scene and point cloud images come from the demo visualization, while the remaining flows and gate-state comparison illustrate the mechanism.
    }
    \label{fig:worldruler-overview}
\end{figure*}

As shown in Fig.~\ref{fig:worldruler-overview}, generation proceeds in consecutive windows. 
For window $n$, the executor evaluates actions $U_n$ against the current state and produces a sequence of object states $\mathcal{S}_n$. 
These states and their spatial bindings form generation conditions $C_n$
\begin{equation}
    \begin{aligned}
        \widehat{V}_n &\sim p_{\theta,\phi}\!\left(
            V_n \mid I_0, H_n, U_n, C_n
        \right),\\
        C_n &= \operatorname{Condition}\!\left(
            \mathcal{G}_n, \mathcal{B}_n, \mathcal{S}_n
        \right),
    \end{aligned}
    \label{eq:worldruler-generation}
\end{equation}
where $H_n$ denotes previously generated visual context, $\theta$ denotes the video generator parameters, and $\phi$ denotes trainable attribute replacement parameters. 
The state sequence follows from actions and rule execution within the current window, without future outcomes from reference videos. 
Object binding identifies where rule targets apply, attribute replacement corrects those regions during denoising, and perception updates correspondences and event evidence for subsequent windows. 
The graph and its joint checkpoint support Rule-State Persistence, while binding and replacement support Rule-to-Visual Realization.

\subsection{Executable 3D Rule Graph and Persistence}
\label{sec:method-graph}

The rule graph organizes object identities, spatial information, and interaction logic in a shared representation. 
Object nodes carry persistent identities, semantic descriptions, 3D support, and attribute states. 
Rule nodes specify triggers, state preconditions, and effect assignments. 
Typed edges distinguish action subjects, action targets, precondition objects, and effect targets, allowing a consequence to occur away from its trigger. 
Shared attributes and event conditions express dependencies between rules. 
For example, collecting a key changes the possession state required by an unlocking rule. 
Initial objects and spatial information are established from the initial observation and available scene geometry. 
References to objects that cannot yet be localized remain in the graph for later binding.

Let $S_t$ collect object attributes and let $q_t$ contain event history, rule activation records, transition progress, and pending effects. 
The execution state is $X_t=(S_t,q_t)$ and evolves according to
\begin{equation}
    X_{t+1}
    = \operatorname{Execute}_{\mathcal{R}}\!\left(
        X_t, u_t, e_t
    \right),
    \label{eq:worldruler-execution}
\end{equation}
where $u_t$ is the current action and $e_t$ is accepted event evidence. 
The executor evaluates action, state, spatial, and event-order conditions before assigning effects to their designated objects. 
Rules requiring action completion consume confirmed completion events, so an action request alone does not satisfy them. 
Unknown attributes remain explicitly unknown and cannot satisfy a precondition through negation.

For changes with a duration, the state records current values, targets, and transition progress. 
Generation receives the effective value at each time, while dependent predicates respect transition completion. 
This prevents a partially opening door from prematurely satisfying an open-door condition. 
Effects on invisible objects persist in the state. When an effect target is unresolved, the system retains its description, rule provenance, and event time until the target can be bound.

A checkpoint jointly serializes the graph, execution state, and bindings
\begin{equation}
    \begin{aligned}
        \mathcal{C}_t
        &= \operatorname{Save}\!\left(\mathcal{G}_t,X_t,\mathcal{B}_t\right),\\
        \operatorname{Load}(\mathcal{C}_t)
        &= \left(\mathcal{G}_t,X_t,\mathcal{B}_t\right).
    \end{aligned}
    \label{eq:worldruler-checkpoint}
\end{equation}
The checkpoint path in Fig.~\ref{fig:worldruler-overview} restores rule definitions, object identities, and execution progress so that subsequent actions continue from the saved conditions. 
For example, a checkpoint taken after defeating a monster but before collecting its key retains the defeated monster, the key on the ground, the closed door, and the remaining collection and unlocking rules. 
Continuing generation additionally requires visual boundary context corresponding to the restored state. 
Images may vary with sampling, while identical subsequent actions and event evidence should produce consistent rule decisions.

\subsection{Dynamic Object Binding and Perceptual Feedback}
\label{sec:method-binding}

Dynamic object binding first identifies which scene instance a rule addresses and then determines where that instance can be controlled in the current view. 
Semantic, appearance, and spatial consistency associate candidate instances with rule entities, and $\mathcal{B}_t$ stores these correspondences. 
Objects of the same category retain distinct identities, while viewpoint changes, occlusion, and reappearance preserve established bindings. 
For localized objects, camera projection and instance segmentation~\cite{ravi2024sam2} provide spatiotemporal masks $M_i$, restricted by visibility and occlusion.

Rules can update an object's persistent state while it is invisible. 
Given an established 3D binding, the stored state is assigned to the projected region before generating the window in which the object first appears. 
A gate opened outside the view is therefore conditioned as open from its first appearance, without repeating the opening action. 
Effects on entirely unlocalized entities remain pending until reliable correspondences become available. 
First-appearance control thus depends on spatial grounding established before generation.

The executed scene geometry guides layout, camera motion, and object motion, while effective states and changes form global and object-level visual descriptions. 
Each local condition specifies an identity, spatiotemporal support, and structured attribute targets. 
Surface changes use object surface support, while motion and deformation cover the current geometry and rule-executed trajectory, excluding foreground occluders and unrelated instances. 
Unknown attributes are omitted from local control, and completed changes persist as resulting-state conditions.

Perception independently estimates identity, visibility, visual attributes, and action completion from generated frames, without receiving desired rule outcomes as answer cues. 
Evidence passing identity, confidence, and temporal validity checks updates bindings, local geometry, and ordinary observable state, and supplies events for subsequent rules that require visual confirmation. 
Rule-prescribed states and observed outcomes are recorded separately. 
If a gate appears closed despite an open-state requirement, the prescribed state is retained, and the observed discrepancy guides correction in subsequent generation windows. 
Feedback affects only windows generated after its acceptance, without rewriting delivered video or treating unconfirmed outcomes as successful.

\subsection{Object Attribute State Replacement}
\label{sec:method-replacement}

\begin{figure*}[t]
    \centering
    \includegraphics[width=\textwidth]{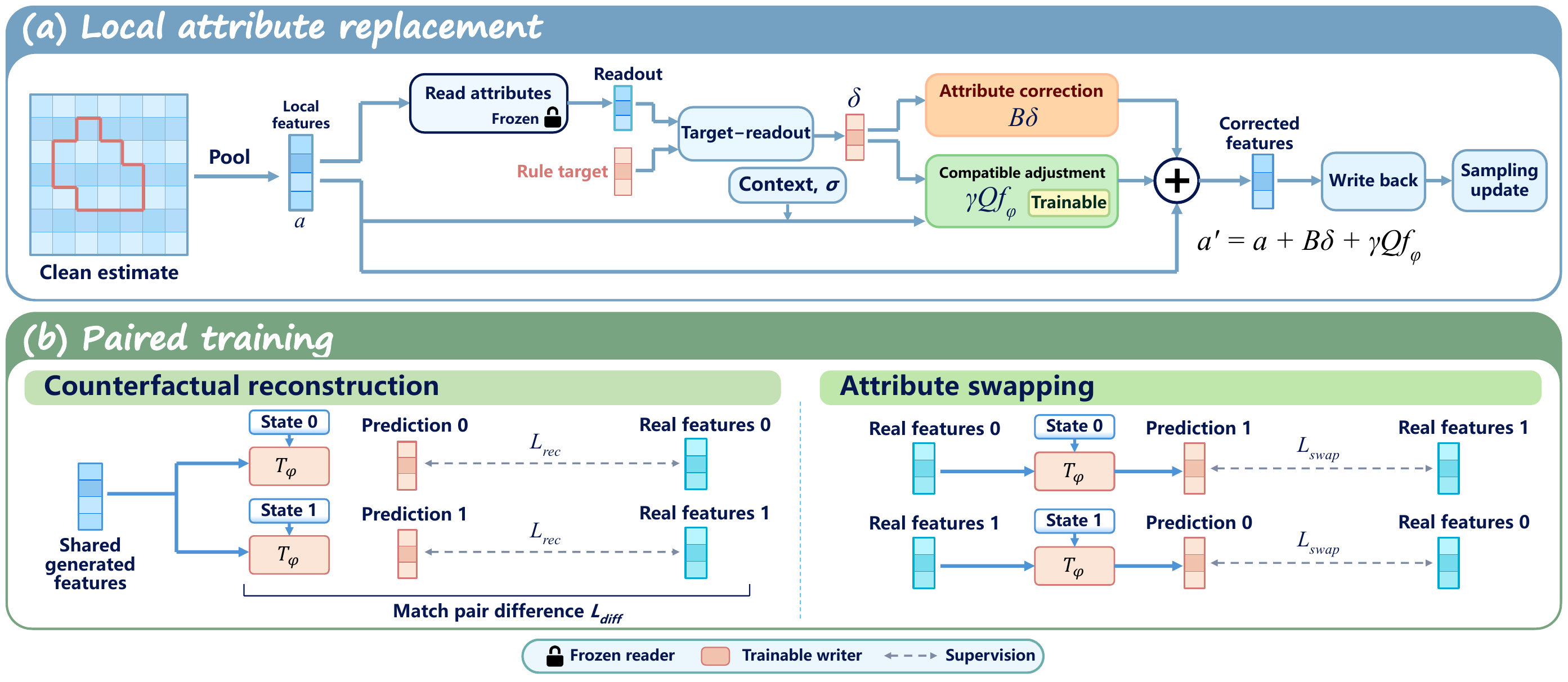}
    \caption{
    Local attribute replacement and paired training. 
    (a) A frozen reader estimates current attributes, and rule targets determine their discrepancies. 
    Attribute corrections and compatible adjustments modify local features before insertion into the sampling update. 
    (b) Counterfactual reconstruction supervises two target states and their difference from shared generated features. 
    Bidirectional swapping starts from real paired features and supervises replacement with the other state. 
    All branches share the replacement operator $T_{\phi}$. Dashed links indicate supervision.
    }
    \label{fig:worldruler-replacement}
\end{figure*}

Object binding determines the support of an edit, and attribute state replacement adjusts its visual representation to the rule target. 
Figure~\ref{fig:worldruler-replacement}(a) expands the Replace module in the overview into local attribute reading, target correction, and insertion into the sampling update.

\paragraph{Reading Local Attributes.}
Let $z_{\sigma}$ denote the video latent at noise level $\sigma>0$ and let $v_{\theta}$ be the predicted velocity. Under the generator's parameterization, the clean estimate is
\begin{equation}
    \widehat{z}_0
    = z_{\sigma} + \sigma v_{\theta}(z_{\sigma},\sigma,C_n).
    \label{eq:worldruler-clean}
\end{equation}
We omit window and latent-frame indices below. Pooling over the object mask yields local features $a_i=\operatorname{Pool}_{M_i}(\widehat{z}_0)$. 
Dynamic regions map to a fixed canonical grid within the object bounds, using only cells with valid spatial support.

A frozen affine reader estimates current attributes from visual features as $\widehat{s}_i=A_i a_i+b_i$. 
It is fitted independently on annotated real training videos and receives no rule targets. 
Rule execution supplies the currently effective target $s_i^{\star}$, yielding the discrepancy $\delta_i=s_i^{\star}-\widehat{s}_i$.

\paragraph{Attribute Correction and Compatible Adjustment.}
We separate direct attribute correction from compatible visual adjustments, as satisfying the attribute readout alone does not guarantee coherent appearance and motion changes.
Here, $A_i$ contains only rows for currently known and controllable attributes. 
When these attributes are independently identifiable and $A_i$ has full row rank, define
\begin{equation}
    B_i=A_i^{\top}(A_iA_i^{\top})^{-1},
    \qquad
    Q_i=I-B_iA_i.
    \label{eq:worldruler-projection}
\end{equation}
The right inverse $B_i$ maps attribute discrepancies into feature space, while $Q_i$ projects additional adjustments into the reader's null space. 
Adding both corrections to the original features defines the local replacement operator $T_{\phi}$

\begin{equation}
    a_i'
    = T_{\phi}(a_i,s_i^{\star})
    = a_i + B_i\delta_i
    + \gamma(\delta_i)\,Q_i
      f_{\phi}\!\left(a_i,h_i,\widehat{s}_i,s_i^{\star},\sigma\right),
    \label{eq:worldruler-replacement}
\end{equation}

where $h_i$ denotes local context, $f_{\phi}$ is a trainable writer, and $\gamma(\delta)=\min(\lVert\delta\rVert_{\infty},c)$ for a fixed cap $c$. 
The term $B_i\delta_i$ corrects attribute coordinates, while the branch constrained by $Q_i$ learns compatible appearance and motion changes without altering the prescribed readout. 
Since $A_iB_i=I$ and $A_iQ_i=0$, full replacement satisfies
\begin{equation}
    A_i a_i'+b_i=s_i^{\star}.
    \label{eq:worldruler-exactness}
\end{equation}
Both corrections vanish when the discrepancy is zero, preserving features that already satisfy the target. 
Multiple active attributes are solved jointly to avoid conflicting sequential writes. 
Equation~\eqref{eq:worldruler-exactness} constrains learned attribute coordinates, while decoded visual semantics and content preservation are evaluated independently on generated videos.

\paragraph{Writing Back to Denoising.}
Local corrections are inserted into the current clean estimate and used to update the sampling velocity. For disjoint object supports,
\begin{equation}
    \begin{aligned}
        \widehat{z}_0'
        &= \widehat{z}_0
        + \sum_i \operatorname{Scatter}_{M_i}(a_i'-a_i),\\
        v_{\theta}'
        &= v_{\theta}
        + \frac{\widehat{z}_0'-\widehat{z}_0}{\sigma}.
    \end{aligned}
    \label{eq:worldruler-scatter}
\end{equation}
The sampler uses $v_{\theta}'$ to advance to the next denoising step, incorporating rule corrections into the generation trajectory. 
Scattering is constructed as a right inverse of pooling over valid canonical cells, and compatible constraints on overlapping supports are solved jointly in their shared latent region. 
Writes exclude fixed history boundaries, invisible regions, and unauthorized attributes. 
Latents outside the support receive no direct correction.

\subsection{Counterfactual Reconstruction and Attribute Swapping}
\label{sec:method-training}

As shown in Fig.~\ref{fig:worldruler-replacement}(b), we train the same replacement operator using video pairs $(V^{(0)},V^{(1)})$ aligned by scene, action context, and rule events.
The branches share scene, camera, and action context while differing in rule-specified attribute states. 
Each uses its own true states and visual outcomes, including changes caused by ordinary actions when the designated rule is disabled. Encoding yields local targets $a^{(0)},a^{(1)}$ with attributes $s^{(0)},s^{(1)}$. 
Unknown, invisible, and padded samples are excluded from the losses.

We first fit a regularized affine reader on real local features and check attribute identifiability. 
We then freeze the reader, video generator, and visual encoder, optimizing only the writer parameters $\phi$. 
Real paired videos provide training supervision, while inference conditions come from the initial observation, generated history, and rule execution.

\paragraph{Counterfactual Reconstruction.}
We extract a local estimate $\bar{a}_{\sigma}$ from a denoising trajectory and supply it to two shared replacement branches with targets $s^{(0)}$ and $s^{(1)}$. 
Write $\widetilde{a}^{(k)}=T_{\phi}(\bar{a}_{\sigma},s^{(k)})$, omitting the shared context and noise level. 
Reconstruction and difference losses are
\begin{equation}
    \begin{aligned}
        \mathcal{L}_{\mathrm{rec}}
        &= \mathbb{E}\!\left[
            \frac{1}{2}\sum_{k\in\{0,1\}}
            \lVert\widetilde{a}^{(k)}-a^{(k)}\rVert_2^2
        \right],\\
        \mathcal{L}_{\mathrm{diff}}
        &= \mathbb{E}\!\left[
            \left\lVert
                (\widetilde{a}^{(1)}-\widetilde{a}^{(0)})
                -(a^{(1)}-a^{(0)})
            \right\rVert_2^2
        \right].
    \end{aligned}
    \label{eq:worldruler-paired-loss}
\end{equation}
Shared inputs require the output difference to arise from the state condition. 
Reconstruction learns each state's visual realization, while difference supervision targets the visual change associated with the attribute change.

\paragraph{Bidirectional Attribute Swapping.}
Swapping starts from one real local representation, writes the other branch's attributes, and reconstructs the corresponding target
\begin{equation}
    \mathcal{L}_{\mathrm{swap}}
    = \mathbb{E}\!\left[
        \frac{1}{2}\sum_{k\in\{0,1\}}
        \left\lVert
            T_{\phi}(a^{(k)},s^{(1-k)})-a^{(1-k)}
        \right\rVert_2^2
    \right].
    \label{eq:worldruler-swap-loss}
\end{equation}
Swapping uses clean inputs at zero noise level and changes only the supervised attribute targets. 
Alignment of unrelated content across the pair lets this objective supervise both attribute changes and local content preservation. 
The total objective is
\begin{equation}
    \mathcal{L}
    = \mathcal{L}_{\mathrm{rec}}
    + \lambda_{\mathrm{diff}}\mathcal{L}_{\mathrm{diff}}
    + \lambda_{\mathrm{swap}}\mathcal{L}_{\mathrm{swap}}.
    \label{eq:worldruler-total-loss}
\end{equation}
All losses are normalized over valid object and temporal support. 
Counterfactual reconstruction learns to realize prescribed states, difference supervision constrains rule-induced visual changes, and attribute swapping encourages compatible adjustments while preserving unrelated content. Local insertion further limits the direct spatial extent of these changes.

\section{RuleVerse}
\label{sec:dataset}

RuleVerse contains paired rule-driven interaction videos collected in Habitat~\cite{puig2024habitat} and GTA-V, covering indoor object interactions and open game environments. 
Each pair shares scene configurations and action intent but differs in whether a designated rule mechanism is enabled, producing distinct state evolutions.
The rule-disabled branch retains ordinary actions and other native reactions, so triggering conditions may still hold. 
Solo samples examine a designated rule, while Full samples cover multi-rule scripts, including explicit dependencies between earlier outcomes and later conditions in Habitat.

\subsection{Data Collection and Annotation}
\label{sec:dataset-collection}

\begin{figure*}[t]
    \centering
    \includegraphics[width=\linewidth]{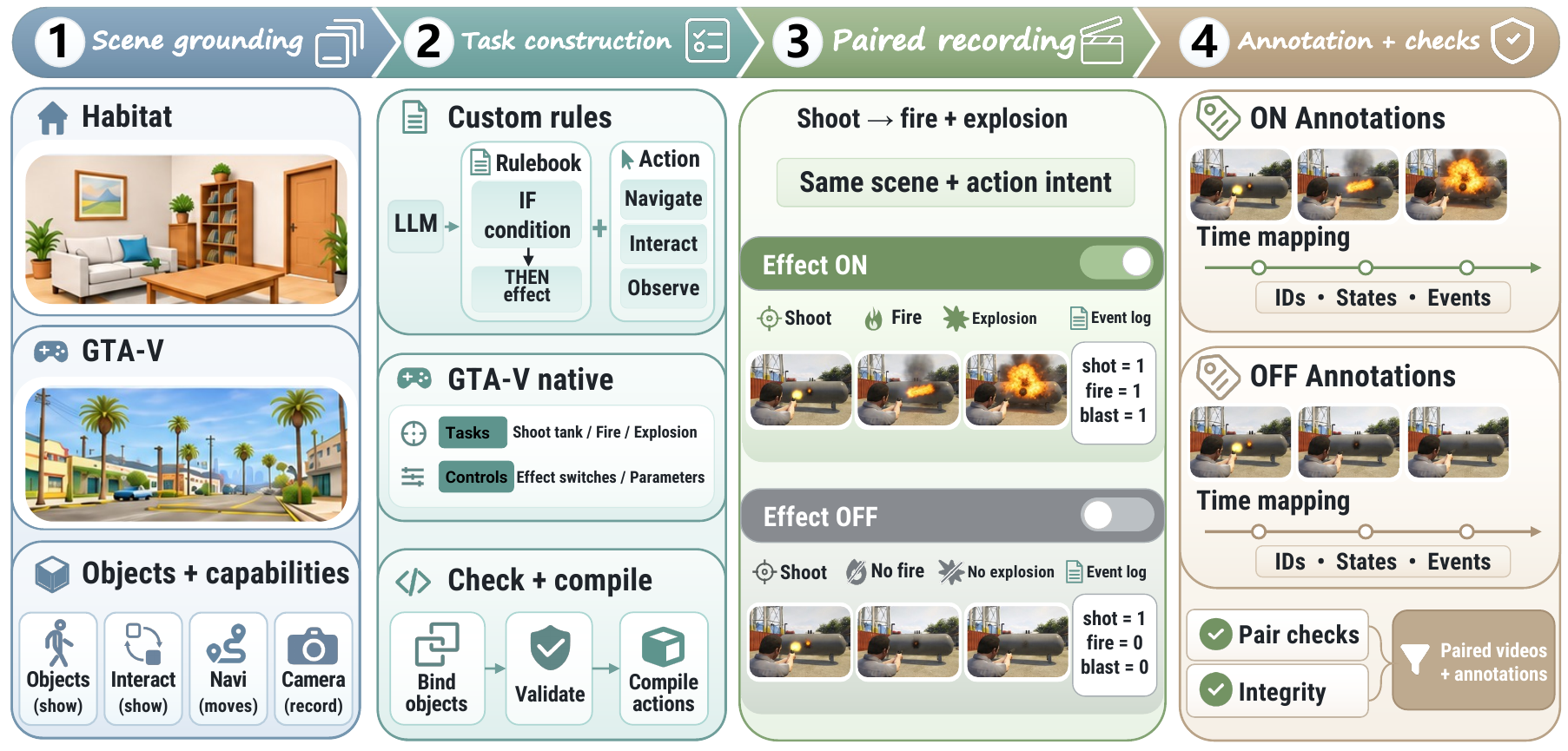}
    \caption{RuleVerse data collection. Scene capabilities constrain rule and script authoring, and compiled actions are executed with the designated rules enabled and disabled. 
    Simulator states and events provide annotations of rule outcomes.
    Videos and annotations are checked before splitting by scene or location group.}
    \label{fig:dataset-pipeline}
\end{figure*}

Simulation-based pipelines enable automated collection of visual data with detailed ground-truth annotations~\cite{greff2022kubric}. 
Figure~\ref{fig:dataset-pipeline} illustrates our data collection process.
We first inspect native objects, parts, spatial relations, and initial states, together with interaction capabilities, reachable paths, and interaction and observation positions. 
Trigger and effect combinations are selected according to scene capabilities, prioritizing underrepresented categories during collection. For custom rules, a language model produces a rulebook containing natural-language descriptions, executable conditions, and state assignments, together with an interaction script. 
Code checks object bindings, syntax, initial conditions, and execution feasibility. 
Habitat collection additionally screens rule quality before recording. 
A compiler translates scripts into navigation, interaction, and camera controls bound to concrete entities.

Both branches execute from their corresponding initial configurations. Habitat reuses the same action and camera control plan. 
GTA-V includes both custom rules and interventions on native mechanisms. 
Disabling custom rules preserves native game reactions, while native interventions suppress only the documented mechanism. 
GTA-V pairs share scene configurations and temporally aligned state and event annotations, while allowing minor variations in character motions due to engine dynamics.

Recording stores videos, object identities, action events, and state trajectories, together with trigger conditions, effect timing, and task outcomes. 
For Habitat, an independent interpreter recomputes conditions and effects from state and event logs, followed by checks of paired controls, video integrity, and annotation correspondence, and deduplication using decoded videos. 
GTA-V retains native execution and observation records, with checks of pair structure and video file identity. 
Habitat annotations align to video frames, while GTA-V preserves the temporal mapping and synchronization uncertainty between screen recordings and engine logs. 
Visibility and event timing distinguish when a rule takes effect from when its target enters the view. 
Public inputs contain initial images, rules, and requested actions, while reference videos and subsequent state annotations remain on the answer side.

\subsection{Dataset Statistics}
\label{sec:dataset-statistics}

The dataset contains 3,200 pairs and 6,400 videos, with 2,900 pairs for training and 300 for testing. 
Habitat is split by base scene, while GTA-V groups samples by source scene and spatial proximity before splitting. 
We characterize the dataset through rule combinations, object binding, activation timing, and dependence on action history.

\begin{figure}
    \centering
    \includegraphics[width=\linewidth]{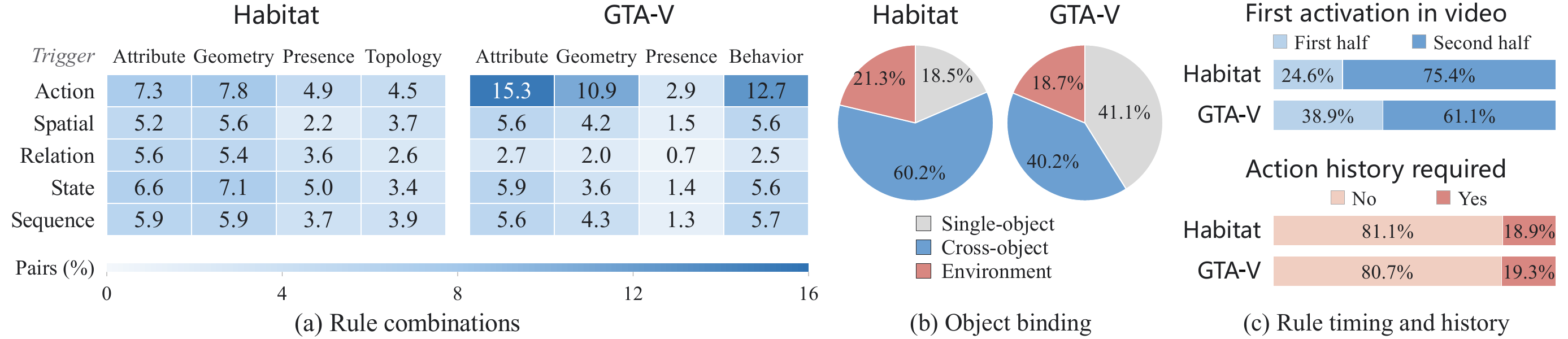}
    \caption{
        Statistics of RuleVerse. 
        \textbf{(a)} Distributions of sample pairs over primary trigger and effect categories, normalized within each displayed heatmap. 
        \textbf{(b)} Rule proportions for single-object binding, cross-object binding, and environment effects.
        \textbf{(c)} First activation in the first or second half of the video (top), and dependence on action history through ordered actions or cumulative event counts (bottom).
    }
    \label{fig:statistics}
\end{figure}

As shown in Figure~\ref{fig:statistics}(a), the heatmaps summarize the joint distribution of each pair’s primary trigger and effect categories. 
The five trigger families cover actions, spatial conditions, relations, states, and sequences. 
The displayed effects include changes in attributes, geometry, and object presence, together with topology changes in Habitat and behavior changes in GTA-V. 
Each heatmap is normalized over the sample pairs within its displayed effect categories. 
Habitat exhibits a more even distribution across these combinations, while GTA-V places greater emphasis on action-triggered effects.

As shown in Figure~\ref{fig:statistics}(b), executed rules are categorized into single-object binding, cross-object binding, and environment effects.
Cross-object rules reference multiple non-actor object or part identities across their conditions and effects, accounting for 60.2\% of Habitat rules and 40.2\% of GTA-V rules. 
Single-object rules involve one such identity, while environment effects modify scene-level properties or routes. 
These categories cover interactions involving individual objects, interactions involving multiple objects, and changes to the surrounding environment.

As shown in Figure~\ref{fig:statistics}(c), the first activation occurs in the second half of the video for 75.4\% of Habitat rules and 61.1\% of GTA-V rules. 
In addition, 18.9\% and 19.3\% of rules, respectively, require action history through ordered actions or cumulative event counts. 
These distributions include cases in which rule conditions must remain available until later interactions and cases whose activation depends on preceding actions.

\subsection{Metrics}
\label{sec:experiment-metrics}
We evaluate rule realization using Rule Success Rate (RSR) and Full Completion Rate (FCR). 
RSR comprises five components evaluated over annotated rule events in rule-enabled videos, while rule-disabled videos provide counterfactual references for evaluating rule-specific changes.
\texttt{Identity} measures whether the relevant trigger objects and effect targets match their reference identities. 
\texttt{Trigger} assesses whether the specified trigger condition is visibly satisfied. 
\texttt{Effect} evaluates whether the prescribed change is realized on the target. 
\texttt{Timing} checks whether the effect begins after the trigger and becomes established within the specified time window. 
\texttt{Hold} measures whether the resulting state persists over the required observation interval.

FCR evaluates complete multi-stage tasks through three components. 
\texttt{Stages} measures the proportion of tasks in which all required rule events are successfully realized, rather than the average success rate of individual stages. 
\texttt{Order} assesses whether all required stages occur in the prescribed chronological order, and \texttt{Goal} checks whether the final task goal is visibly achieved at the end of the video. The average FCR is the arithmetic mean of these three independently evaluated components.

Visual judgments are obtained using a frozen vision-language evaluator with fixed-choice questions and initial identity references. 
Generated videos are sampled at 3 fps, including the first and last frames. 
The same judging protocol is applied to all compared methods. Uncertain or invalid judgments remain in the corresponding denominators and receive no success credit. 
All scores are reported as percentages.
\section{Experiments}
\label{sec:experiments}

Our experiments evaluate WorldRuler along two dimensions: Rule-State Persistence and Rule-to-Visual Realization. 
We first compare rule realization and task-level performance on RuleVerse, followed by an external evaluation on WorldState-Bench. 
We then examine dynamic object binding, attribute state replacement, and the swapping objective through component ablations, and evaluate perceptual and visual consistency after saving and loading.
Qualitative examples illustrate how the runtime produces different outcomes under enabled and disabled rule mechanisms.

\subsection{Implementation}
\label{sec:experimental-setup}

We use the MiniMax-H3 Ref2VA Turbo for video generation at a resolution of $1344\times768$ and 24 fps.
Each generation window contains 124 frames. 
We use the native H3 Qwen encoder for generation conditioning, along with Qwen3.8-Max~\cite{qwen38}, SAM2~\cite{ravi2024sam2}, and DA3~\cite{depthanything3} for visual observations, instance segmentation, and geometry estimation.

Attribute state replacement is applied to local clean estimates at each denoising step, with corrections written back before the next sampling update.
Following Sec.~\ref{sec:method-training}, we first fit the affine attribute reader, then freeze the reader, video generator, and visual encoder while optimizing only the attribute writer. 
Training uses paired reconstruction, paired difference, and bidirectional attribute swapping losses, excluding unknown attributes, invisible regions, and padded frames.

\subsection{Main Results}
\label{sec:main-results}
\label{sec:external-experiments}

\paragraph{\dataname{}.}
As shown in Table~\ref{tab:main-results}, WorldRuler outperforms the compared baselines across both the Habitat and GTA-V subsets.
On the Habitat subset, it achieves average RSR and FCR scores of 64.99\% and 48.04\%, exceeding SWM by 20.93 and 30.06 percentage points, respectively. 
Compared with SWM, effect accuracy increases from 20.74\% to 58.47\%, timing accuracy from 15.67\% to 37.36\%, and hold accuracy from 17.05\% to 50.83\%. 
WorldRuler also improves all three task-level components, reaching 52.61\% for stages, 44.38\% for order, and 47.12\% for goal. 
These results support the effectiveness of combining explicit rule execution with state-conditioned generation for realizing rule effects and dependent interactions.

On the GTA-V subset, WorldRuler achieves an average RSR of 67.42\%, compared with 44.19\% for SWM and 43.43\% for MiniMax-H3. 
It obtains 60.00\% for stages, 60.00\% for order, and 80.00\% for goal, yielding the highest average FCR of 66.66\%. 
The improvements across \dataname{} indicate that the framework benefits indoor object interactions as well as open game environments. 

\begin{table*}[t]
    \centering
    \small
    \caption{
    Comparison on the Habitat and GTA-V subsets of \dataname{}.
    We report component-wise and average Rule Success Rate (RSR) and Full Completion Rate (FCR). 
    All values are percentages.
    }
    \label{tab:main-results}
    \begin{tabular}{lcccccccccc}
        \toprule
        Method & \multicolumn{6}{c}{Rule Success Rate $\uparrow$} & \multicolumn{4}{c}{Full Completion Rate $\uparrow$} \\
        \cmidrule(lr){2-7}
        \cmidrule(lr){8-11}
         & Identity & Trigger & Effect & Timing & Hold & avg. & Stages & Order & Goal & avg. \\
        \midrule
        \multicolumn{7}{l}{\textit{Habitat}} \\
        MiniMax-H3~\cite{minimax2026h3} & 78.85 & 53.37 & 17.79 & 9.13 & 9.62 & 33.75 & 11.21 & 13.64 & 5.26 &  10.03 \\		
        CWM~\cite{chen2026codeworldmodelcoding} & 93.85 & 49.23 & 3.85 & 0.77 & 0.77 & 29.69 & 13.52 & 12.48 & 16.81 & 14.27 \\
        SWM~\cite{ki2026stateworldmodelsexecute} & 98.16 & 68.66 & 20.74 & 15.67 & 17.05 & 44.06 & 15.55 & 17.58 & 20.82 & 17.98 \\
        \textbf{WorldRuler} & \textbf{98.61} & \textbf{79.72} & \textbf{58.47} & \textbf{37.36} & \textbf{50.83} & \textbf{64.99} & \textbf{52.61} & \textbf{44.38} & \textbf{47.12} & \textbf{48.04} \\

        \midrule
        \multicolumn{7}{l}{\textit{GTA-V}} \\
        MiniMax-H3~\cite{minimax2026h3} & 54.76 & 60.48 & 34.76 & 34.29 & 32.86 & 43.43 & 20.00 & 0.00 & 20.00 & 13.33 \\
        CWM~\cite{chen2026codeworldmodelcoding} & 53.33 & 46.67 & 22.86 & 11.43 & 22.86 & 31.43 & 0.00 & 0.00 & 20.00 & 6.67 \\
        SWM~\cite{ki2026stateworldmodelsexecute} & 69.52 & 64.76 & 31.43 & 24.76 & 30.48 & 44.19 & 0.00 & 0.00 & 0.00 & 0.00 \\
        \textbf{WorldRuler} & \textbf{92.73} & \textbf{80.02} & \textbf{51.26} & \textbf{44.49} & \textbf{68.62} & \textbf{67.42} & \textbf{60.00} & \textbf{60.00} & \textbf{80.00} & \textbf{66.66} \\
        \bottomrule
    \end{tabular}
\end{table*}

\paragraph{WorldState-Bench.}
We evaluate the complete WorldRuler framework on the 50 official WorldState-Bench~\cite{ki2026stateworldmodelsexecute} pairs using the benchmark's original judging protocol.
This configuration uses a single 124-frame window, resolves the submitted action and rules before generation, and permits up to two feedback corrections before publishing the clip.
The evaluation includes the trained attribute state replacement module alongside rule execution, dynamic object binding, and perceptual feedback.

As shown in Table~\ref{tab:worldstate-results}, WorldRuler matches SWM on EndFollow, TwinContrast, and Strict, while achieving higher scores on PairFollow, CntAbsent, and PairStrict.
These three scores reach 94\%, 99\%, and 92\%, compared with 92\%, 98\%, and 90\% for SWM.
WorldRuler improves PairFollow and PairStrict by two percentage points each over SWM.
These results indicate competitive performance on an external rule-following benchmark, with modest gains over SWM under the reported configurations.

\begin{table*}[t]
    \centering
    \small
    \caption{Evaluation of WorldRuler and baseline methods on WorldState-Bench using the benchmark's original judging protocol.}
    \label{tab:worldstate-results}
    \begin{tabular}{lcccccc}
        \toprule
        Method & EndFollow $\uparrow$ & PairFollow $\uparrow$ & TwinContrast $\uparrow$ & Strict $\uparrow$ & CntAbsent $\uparrow$ & PairStrict $\uparrow$ \\
        \midrule
        Wan2.2 A14B~\cite{wan2025} & 65 & 32 & 38 & 54 & 63 & 26 \\
        HunyuanVideo 1.5~\cite{wu2025hunyuanvideo} & 57 & 20 & 26 & 29 & 30 & 4 \\
        MiniMax-H3~\cite{minimax2026h3} & 66 & 32 & 32 & 29 & 30 & 4 \\
        LingBot-World 2.0~\cite{gao2026infinite} & 48 & 16 & 38 & 44 & 76 & 14 \\
        AlayaWorld~\cite{team2026alayaworldv11} & 37 & 12 & 28 & 33 & 85 & 6 \\
        SWM~\cite{ki2026stateworldmodelsexecute} & 96 & 92 & 96 & 95 & 98 & 90 \\
        \textbf{WorldRuler} & \textbf{96} & \textbf{94} & \textbf{96} & \textbf{95} & \textbf{99} & \textbf{92} \\
        \bottomrule
    \end{tabular}
\end{table*}

\subsection{Ablation Study}
\label{sec:ablation-experiments}

We conduct ablation studies on the Habitat subset of \dataname{} to examine the contributions of dynamic object binding, attribute state replacement, and the swapping objective, as well as the information required for save/load persistence. 

\paragraph{Component Ablation.}
We examine dynamic object binding, attribute state replacement, and the swapping objective on the Habitat subset of \dataname{}.
Alongside RSR and FCR, we evaluate First-appearance and Preservation. 
First-appearance measures whether targets affected by rules while off-screen exhibit the correct identities and prescribed attributes upon entering the view and during the immediately following observation interval. 
Preservation measures whether unrelated objects and attributes remain intact, excluding changes permitted by ordinary actions or the rules.

As shown in Table~\ref{tab:ablations}, the full method achieves the highest scores across all four metrics. 
Replacing dynamic object binding with fixed binding reduces average RSR from 64.99\% to 48.09\% and First-appearance from 29.69\% to 20.58\%, supporting the importance of updating object correspondences when applying persistent rule states to visual instances. 
Removing attribute state replacement reduces average RSR and FCR by 18.56 and 9.54 percentage points, respectively, while also lowering First-appearance and Preservation. 
These results support local attribute replacement as a contributor to rule realization and content preservation.

Removing the swapping objective reduces average RSR to 51.24\%, First-appearance to 6.45\%, and Preservation to 20.63\%. 
Its first-appearance score is also lower than that obtained without attribute replacement, suggesting that the effectiveness of the learned replacement module depends on its training objective. 
The full method improves both first-appearance correctness and preservation.

\begin{table*}[t]
    \centering
    \small
    \caption{
    Component ablations on the Habitat subset of \dataname{}. All configurations use the complete test set, with rule realization evaluated on rule-enabled branches and rule-disabled branches retained for counterfactual comparison.
    }
    \label{tab:ablations}
    \begin{tabular}{lcccc}
        \toprule
        Variant & RSR $\uparrow$ & FCR $\uparrow$ & First-appearance $\uparrow$ & Preservation $\uparrow$ \\
        \midrule
        Fixed object binding & 48.09 & 44.81 & 20.58 & 19.49 \\
        Without attribute state replacement & 46.43 & 38.50 & 12.90 & 17.16 \\
        Without $\mathcal{L}_{\mathrm{swap}}$ & 51.24 & 45.75 & 6.45 & 20.63 \\
        \textbf{Full Method} & \textbf{64.99} & \textbf{48.04} & \textbf{29.69} & \textbf{28.71} \\
        \bottomrule
    \end{tabular}
\end{table*}

\paragraph{Save and Load.}
\label{sec:persistence-experiments}
We compare three restoration settings, namely Last image and rule text, Rules and attributes without execution progress, and Joint rules, execution state, and bindings. 
All settings use identical checkpoints and continuation actions to examine how the retained information affects interaction after restoration.

Consistency evaluates whether post-load perceptual observations, visible object states, and generated scene content remain consistent with the saved interaction state and prescribed rule outcomes. 
It is assessed from the resumed generation. 
Post-load RSR evaluates the visual realization of pending and subsequent rule events, while Post-load FCR evaluates stages, order, and goal performance for the continuation tasks using the metric definitions in Section~\ref{sec:experiment-metrics}.

As shown in Table~\ref{tab:persistence}, jointly preserving rules and execution state achieves the highest Consistency of 98.17\%, compared with 94.25\% when restoring from the last image and rule text and 95.13\% when execution progress is omitted. 
Relative to the latter setting, the joint checkpoint also improves Post-load RSR from 51.26\% to 54.36\% and Post-load FCR from 45.53\% to 47.89\%. 
These results support preserving execution progress alongside rules and attributes to maintain perceptual and visual consistency and improve subsequent rule-governed interaction.

\begin{table}[t]
    \centering
    \small
    \caption{
    Save and load evaluation. 
    Consistency uses all predetermined checkpoints, while post-load metrics use predetermined continuation events and tasks.
    }
    \label{tab:persistence}
    \begin{tabular}{lccc}
        \toprule
        Restoration & Consistency $\uparrow$ & Post-load RSR $\uparrow$ & Post-load FCR $\uparrow$ \\
        \midrule
        Last image and rule text & 94.25 & 49.23 & 40.11 \\
        Rules and attributes, no progress & 95.13 & 51.26 & 45.53 \\
        \textbf{Joint rules, execution state, and bindings} & \textbf{98.17} & \textbf{54.36} & \textbf{47.89} \\
        \bottomrule
    \end{tabular}
\end{table}

\subsection{Qualitative Results}
\label{sec:realization-experiments}

\begin{figure*}[t]
    \centering
    \includegraphics[width=\textwidth]{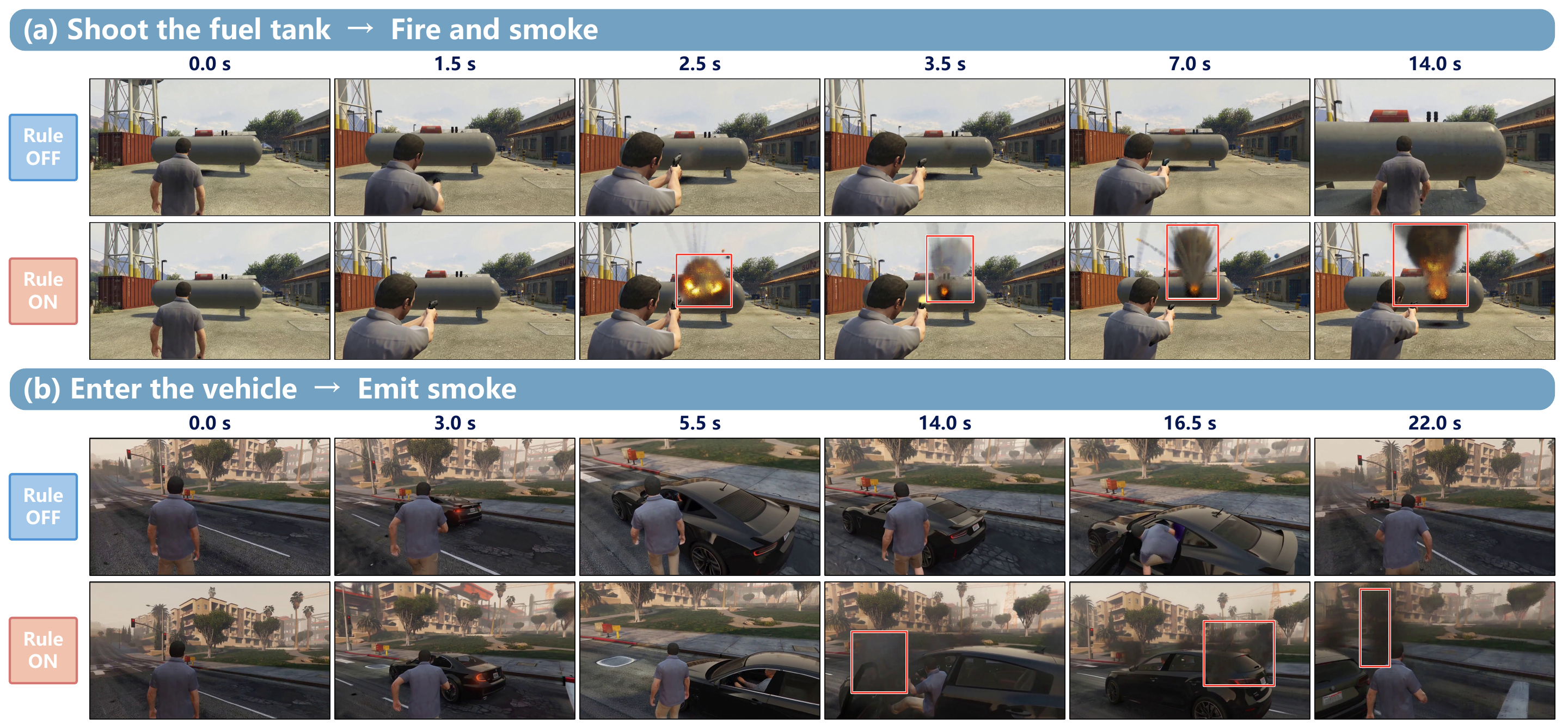}
    \caption{Qualitative results on the GTA-V subset of RuleVerse from the WorldRuler runtime with MiniMax-H3 Ref2VA. The two cases show fire and smoke after shooting a fuel tank, and smoke after entering a vehicle. Enabled and disabled branches are displayed at matching video times, with red boxes marking visible effect regions.}
    \label{fig:experiment-qualitative}
\end{figure*}

Qualitative comparisons use consecutive segments around rule effects, aligned by action stage, video time, and object identity. 
Figure~\ref{fig:experiment-qualitative} presents two cases involving shooting a fuel tank to produce fire and smoke and entering a vehicle to trigger smoke. 
Rule-enabled and rule-disabled branches are shown at matching video times, with red boxes highlighting the affected regions. 
These comparisons illustrate how the runtime generates different visual outcomes under enabled and disabled rule mechanisms.

\section{Conclusion}
\label{sec:conclusion}


We presented WorldRuler, a framework for interactive video world models governed by user-defined rules. An executable 3D rule graph jointly maintains rule definitions, dependencies, and execution states to support saving and restoration. 
Dynamic object binding and perceptual feedback connect persistent states to visual instances across generation windows, while attribute state replacement with counterfactual reconstruction and swapping provides local control over rule-prescribed changes. 
We also introduced RuleVerse, a dataset of paired rule-enabled and rule-disabled interactions with detailed state and event annotations.
Experiments show improved rule realization and task-level performance across Habitat and GTA-V, together with high perceptual and visual consistency after restoration. 

\bibliographystyle{abbrv}
\bibliography{references}

\end{document}